\documentclass[10pt,twocolumn,letterpaper]{article}

\usepackage{cvpr}              

\usepackage[dvipsnames]{xcolor}

\definecolor{cvprblue}{rgb}{0.21,0.49,0.74}
\usepackage[pagebackref,breaklinks,colorlinks,citecolor=cvprblue]{hyperref}

\usepackage{xcolor}

\usepackage{balance}

\usepackage{amsmath,amsfonts}
\usepackage{amssymb}

\usepackage{siunitx}

\definecolor{olivegreen}{rgb}{0, 0.6, 0}
\definecolor{redorange}{HTML}{FF5349}
\definecolor{blue(ncs)}{rgb}{0.0, 0.53, 0.74}
\definecolor{navy}{HTML}{273BE2}

\definecolor{black}{HTML}{000000}
\definecolor{white}{HTML}{ffffff}
\definecolor{color1}{HTML}{ACE5EE}
\definecolor{color2}{HTML}{0093AF}
\definecolor{color3}{HTML}{CC0000}
\definecolor{color4}{HTML}{0087BD}
\definecolor{color5}{HTML}{333399}
\definecolor{color6}{HTML}{20B2AA}

\usepackage{xspace}

\usepackage{graphicx}
\usepackage[export]{adjustbox}
\usepackage{wrapfig}

\usepackage{enumitem} 
\usepackage{tcolorbox}

\usepackage{verbatim}

\usepackage{booktabs}
\usepackage{multirow}
\usepackage{makecell}

\usepackage{nicefrac}

\usepackage{algorithm}
\usepackage{algpseudocode}

\usepackage{caption}
\usepackage{subcaption}
\usepackage{amsmath}
\DeclareMathOperator*{\argmax}{arg\,max}

\usepackage{bbm}
\usepackage{arydshln}

\usepackage{pifont}
\usepackage{txfonts}
\usepackage{latexsym}

\usepackage[accsupp]{axessibility} 

\newcommand{\thiswork}{PeerAiD\xspace}
\newcommand{\Thiswork}{PeerAiD\xspace}
\newcommand{\scheme}{peer tutoring\xspace}
\newcommand{\Scheme}{Peer tutoring\xspace}
\newcommand{\SCHEME}{Peer Tutoring\xspace}

\newcommand{\JL}[1]{{\color{cyan}[\textbf{\sc JLee}: \textit{#1}]}}
\newcommand{\JW}[1]{{\color{orange}[\textbf{\sc JJung}: \textit{#1}]}}
\newcommand{\JY}[1]{{\color{blue(ncs)}[\textbf{\sc JSong}: \textit{#1}]}}
\newcommand{\HS}[1]{{\color{navy}[\textbf{\sc HJang}: \textit{#1}]}}

\usepackage{ifthen}
\newboolean{sepline} 
\setboolean{sepline}{true} 

\iffalse 
\renewcommand{\JL}[1]{}
\renewcommand{\JW}[1]{}
\renewcommand{\JY}[1]{}
\renewcommand{\HS}[1]{}
\else
\fi

\usepackage{tikz}

\def\paperID{9540} 
\def\confName{CVPR}
\def\confYear{2024}

\title{\thiswork: Improving Adversarial Distillation from a Specialized Peer Tutor}

\author{Jaewon Jung, Hongsun Jang, Jaeyong Song, and Jinho Lee\thanks{Corresponding author}\\
Department of Electrical and Computer Engineering, Seoul National University, Seoul, Korea\\
{\tt\small \{jungjaewon, hongsun.jang, jaeyong.song, leejinho\}@snu.ac.kr}
}

\begin{document}
\maketitle
\begin{abstract}
Adversarial robustness of the neural network is a significant concern when it is applied to security-critical domains.
In this situation, adversarial distillation is a promising option which aims to distill the robustness of the teacher network to improve the robustness of a small student network.
Previous works pretrain the teacher network to make it robust against the adversarial examples aimed at itself.
However, the adversarial examples are dependent on the parameters of the target network.
The fixed teacher network inevitably degrades its robustness against the unseen transferred adversarial examples which target the parameters of the student network in the adversarial distillation process.
We propose \thiswork to make a peer network learn the adversarial examples of the student network instead of adversarial examples aimed at itself.
\thiswork is an adversarial distillation that trains the peer network and the student network simultaneously in order to specialize the peer network for defending the student network.
We observe that such peer networks surpass the robustness of the pretrained robust teacher model against adversarial examples aimed at the student network.
With this peer network and adversarial distillation, \thiswork achieves significantly higher robustness of the student network with AutoAttack (AA) accuracy by up to 1.66$\%p$ and improves the natural accuracy of the student network by up to 4.72$\%p$ with ResNet-18 on TinyImageNet dataset.
Code is available at \url{https://github.com/jaewonalive/PeerAiD}.
\vspace{-10pt}

\end{abstract}    
\section{Introduction}
\label{sec:intro}

\begin{figure}[t]
\centering
        \includegraphics[width=0.95\columnwidth]{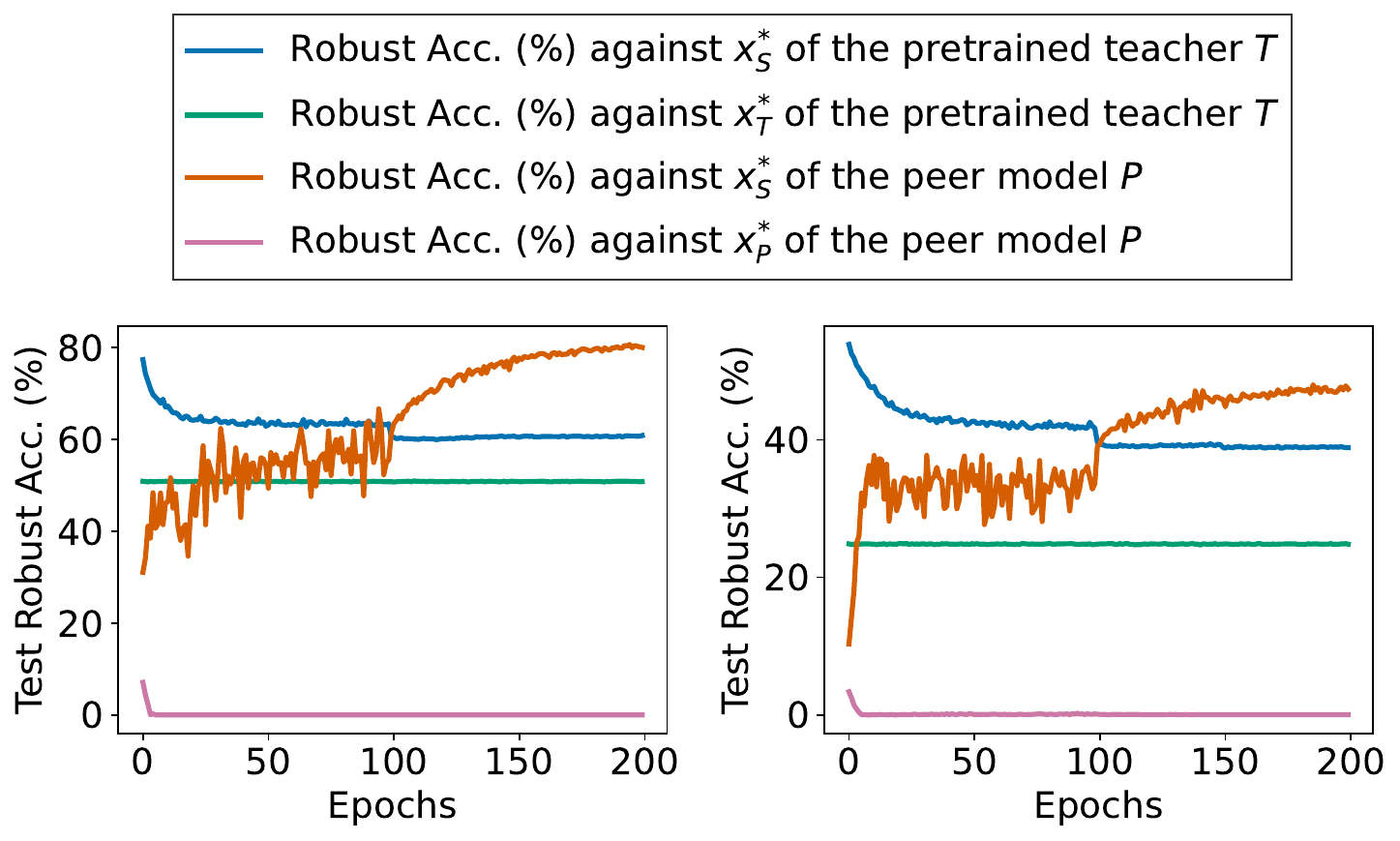}
        \caption{Test robust accuracy of a pretrained robust teacher and a peer network trained from scratch. $x^{*}_{S}$, $x^{*}_{T}$ and $x^{*}_{P}$ denotes the adversarial examples generated from the student, teacher, and peer model, respectively.
        Results on CIFAR-10 (left) and CIFAR-100 (right) show that the peer sustains increasing robustness for the student not provided by the pretrained teacher.
        \vspace{-10pt}}
        \label{fig:intro}
\end{figure}

Deep learning is undoubtedly an irreplaceable tool in many domains such as images~\cite{resnet, vit, swin}, natural language processing~\cite{gpt,bert,llama2}, voice recognition~\cite{speech,wav2vec}, and many real-life scenarios.
However, it has been found that DNNs are vulnerable to an imperceptible noise crafted by attackers~\cite{fgsm, pgd}, and this severely raises concerns about deploying DNNs in critical domains, e.g., autonomous driving~\cite{auto_car} and healthcare~\cite{health}.
Though~\cite{pgd} found that a large model has better robustness than a smaller neural network, a large model is not always applicable to all circumstances, especially in an edge device which has a small size of memory and limited computational capability.
Currently, the only defense technique known to be effective is a variant of adversarial training~\cite{fgsm,pgd}, which essentially trains a model from attacked samples to gain robustness.
Even though the training samples are adversarially perturbed, guiding the training with the correct label allows the model to learn features that are not fooled by similar perturbations.

Among the adversarial training family, one promising and popular way to enhance robustness of the small student network is adversarial distillation (AD), which uses a robustly pre-trained teacher to guide the student. 
In an extenstion of knowledge distillation~\cite{distilling_hinton} which uses the pretrained teacher to approximate the label distribution of the data,
many AD methods~\cite{iad, rslad, adaad} use the pretrained robust teacher network which approximates the label distribution of adversarial examples aimed at the student network.

For such AD methods, an underlying assumption is that the robustness contained within the pretrained teacher is maintained along the adversarial distillation process.
In other words, we expect the teacher to provide a defense for the adversarial examples produced by the student model during adversarial training.
However, such an assumption does not hold in adversarial distillation, especially as the student is trained for several epochs toward convergence.

In \cref{fig:intro}, we test if such an assumption holds in actual training. 
Over 200 epochs of AD training, we attack the student network to create perturbed samples using Projected Gradient Descent (PGD)~\cite{pgd}. 
With the blue curves, we plot the robust accuracy the teacher achieves against those samples (attacked with students).
Initially, the teacher provides a good defense against the student-attacked samples, whose prediction could be a reliable guide to the student \cite{iad}.
However, the robustness quickly drops as the student is trained, and further reduces at the learning rate step.
In fact, similar phenomena have been reported in other literature~\cite{iad}. 
IAD~\cite{iad} partially trusts the teacher network depending on its reliability. 
AKD$^{2}$~\cite{akd} uses the naturally trained teacher together with the pre-trained robust teacher.
MTARD~\cite{MTARD} also employs both teachers, balancing their influence based on how much students converge towards each teacher.
However, while these approaches are effective to some degree, they are limited in that they do not improve the guidance of the teacher, but only reduce the effect of some bad guidance.

In such a regard, the orange curves in \cref{fig:intro} show an intriguing observation.
We use the same teacher model but is randomly initialized and trained from scratch as a \textit{peer} against the student-attacked samples. 
While it is within the expectation that the robust accuracy goes up, the robust accuracy against transferred adversarial examples $x^{*}_{S}$ from the student network reaches much higher than that of the robustly pretrained teacher.
However, the peer has almost no defense (close to 0\%) against adversarial samples $x^{*}_{P}$ that attack itself (i.e., peer-attacked samples).
This states that the peer is specialized at defending against attacks on a student, instead of being a general robust model.

From these, we propose \thiswork, a new AD method that achieves much higher adversarial robustness from training a peer tutor of the target student model.
The method contains the structure to train a peer model for AD, in addition to a novel loss function to train the student to take better guidance from the peer.
Our contributions are summarized as follows:
\begin{itemize}
    \item We observe that training a peer model from the student-attacked sample can build a peer tutor with better guidance for adversarial distillation.
    \item We propose \thiswork that trains a peer using adversarial examples aimed at the student and uses it for AD.
    \item We propose a loss function that is suitable for peer-tutored adversarial distillation.
    \item An extensive set of experiments show that \thiswork gains significantly higher robust accuracy over the prior art in several models and datasets.
\end{itemize}

\section{Related Work}
\label{sec:related work}

\begin{figure*}
     \centering
         \includegraphics[width=0.9\textwidth]{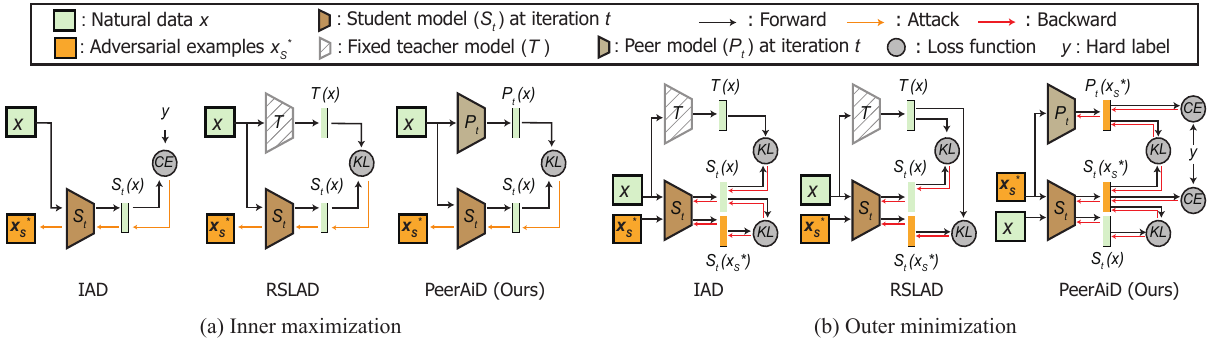}
         \caption{ The adversarial distillation procedure overview of baselines~\cite{iad,rslad} and \thiswork. (a) describes inner maximization to generate adversarial examples from the student model. Baselines use hard labels or pretrained teachers for this. On the other hand, \thiswork uses a peer model. (b) illustrates the outer minimization procedure to optimize model parameters. Baselines only train the student models with the prediction of the pre-trained teachers, but \thiswork trains both peer and student models simultaneously.
         \vspace{-5pt}}
         \label{fig:overview}
\end{figure*}

\subsection{Adversarial Training}
Adversarial Training (AT)~\cite{fgsm, pgd} is an effective method that defends against many white-box and black-box attacks ~\cite{fgsm,pgd,autoattack}. 
Adversarial training is a robust optimization problem and consists of inner maximization and outer minimization.
Mathematically, adversarial training can be formulated as follows. 
\begin{align}\label{at_eqn}
\underset{\theta}{\min}L_{min}(f(\theta, x^{*}_{i}), y_{i}) 
\end{align}
\centerline{where $x^{*}_{i} = \underset{\tilde{x}_{i}\in B(x_{i},\epsilon)}{\argmax}L_{max}(f(\theta,\tilde{x}_{i}), y_{i})$}
$\{(x_{i},y_{i})\}_{i=1}^{N}$ is a training dataset with $N$ samples of input $x_{i}$ and label $y_{i}$.
$f$ is a neural network with the parameter $\theta$.
$L_{max}$ is a loss which is used to find the adversarial examples which increases the loss of the neural network.
The parameter of the neural network is optimized by the $L_{min}$ toward the direction which reduces this $L_{min}$.
$B(x_{i},\epsilon)$ is a ball which restricts the distance between the adversarial example $\tilde{x}_{i}$ and the original data $x_{i}$. 
Popularly used constraint involves $l^{\infty}$ norm and $B(x_{i},\epsilon) = \{\tilde{x}_{i}  \mid  \lVert x_{i}-\tilde{x}_{i} \rVert_{\infty} \leq \epsilon\}$ is widely used in adversarial training.
FGSM~\cite{fgsm} proposes a single-step attack that uses only one iteration of getting gradients of inputs and taking the sign of the gradients to solve the inner maximization.
PGD~\cite{pgd} was developed to iteratively solve the inner maximization problem.
After that, many works~\cite{trades, rslad} adopted Kullback-Leibler (KL) divergence loss in inner maximization to find a better solution for the inner maximization problem to improve the quality of adversarial examples.
The outer minimization problem is to find model parameters that reduce classification loss (e.g., Cross Entropy ) given adversarial examples.
The model parameters solved by outer minimization in adversarial training give higher robust accuracy than the model parameters found by standard training.

\subsection{Adversarial Distillation}

Many researchers studied adversarial distillation to transfer the robustness of a large teacher network to the student network. 
ARD~\cite{ard} adopts the idea of standard knowledge distillation.
It used the prediction of the robust teacher network on natural data to guide the student network.
AKD$^{2}$\cite{akd} show that adversarial distillation and weight averaging~\cite{adv_swa} could prevent robust overfitting problem~\cite{double_descent, gradalign}.
IAD~\cite{iad} focuses on the reliability of teacher networks during adversarial distillation. 
It claims that adversarial samples generated from a student network become challenging in later epochs, so this makes the student network more trustable itself, while the teacher becomes more unreliable on adversarial training data generated from the student network.
RSLAD~\cite{rslad} finds that the inner maximization process of generating adversarial samples could be improved by replacing the hard label with the soft label produced by a robust teacher network on the natural data.
AdaAD~\cite{adaad} presents that maximizing prediction discrepancy between a student network and a robust teacher network could improve the inner maximization process of adversarial training.
CAT~\cite{cat} proposes to simultaneously train multiple robust student networks and exchange their adversarial examples among themselves.
CAT aligns with online knowledge distillation~\cite{dml, guo_online, niyaz, wu2021peer} which trains multiple student networks simultenously in standard training.

\section{Proposed Method}

\subsection{Preliminary}

Adversarial distillation usually pretrains the robust teacher with the adversarial training following \cref{at_eqn}.
The pretrained teacher $T(\cdot)$ is used to produce soft label with natural data $x_{i}$ or adversarial examples $ x^{*}_{i, S}$ of a student model $S(\cdot)$. 
In adversarial distillation, $T(x_{i})$ and $T(x^{*}_{i, S})$ replace the hard label $y_{i}$ in the outer minimization of \cref{at_eqn}.
The KL divergence loss is popularly chosen as $L_{min}$.
RSLAD~\cite{rslad} found that replacing the hard label $y_{i}$ with the soft label produced by the pretrained robust teacher in the inner maximization improves the robust accuracy in adversarial training. 
It regards a prediction of the pretrained robust teacher on natural data as a fixed target and chose $KL(T(x_{i})||S(\tilde{x}_{i}))$ as $L_{max}$.

\subsection{\SCHEME}

We suggest \emph{\scheme}  to use online knowledge distillation in an adversarial distillation setting.
\Scheme trains the peer model $P$ with the student model $S$ from scratch while making the peer model robust to the adversarial examples aimed at the student model.
The peer model provides reliable guidance with \scheme because it becomes much more robust to the adversarial examples generated from the student model.

\textbf{Inner maximization.}
As illustrated in \cref{fig:overview}(a), previous approaches for inner maximization can be categorized into two ways.
The former 
~\cite{iad, akd}
uses the hard label $y$ only.
The latter 
~\cite{rslad, adaad}
uses the prediction of the adversarily pre-trained teacher to generate adversarial examples of the student network.
However, each approach has an obvious limitation.
First, using only hard labels loses the chance to learn the probability distribution of non-target classes from other networks. 
Second, the fixed pre-trained robust teacher network has a limitation on the natural accuracy, which indicates how well the teacher network approximates the true label distribution of the natural samples $x$. 
This trade-off between robustness and the natural accuracy is theoretically studied \cite{trades, tradeoff_odds} and empirically observed phenomenon \cite{reconcile} in many works.

Instead, \thiswork uses the prediction of a peer network which interactively 
learns with the student network to generate adversarial examples $x^{*}_{S}$ of the student network $S$ as depicted in \cref{fig:overview}(a).
The adversarial examples $x^{*}_{S}$ are generated with PGD~\cite{pgd} by finding the gradients of input which increases the KL divergence between the prediction of the peer model and the student model on the training data.
With \scheme, the peer model provides the approximated label distribution to the student model. 
The label distribution provides information on non-target classes which is not contained in the hard label.
The peer model also does not suffer from the degradation in the natural accuracy, which will be discussed in \cref{sec:exp_outer_min}.

\textbf{Outer minimization.} 
Previous approaches~\cite{iad, rslad, adaad} used the prediction of pre-trained teachers to distill better robustness. 
However, the teacher models are adversarially pretrained with adversarial examples aimed at themselves. 
The teacher model has never seen adversarial examples of the student network during the pretraining process.

On the other hand, \thiswork trains the peer network and the student network simultaneously with the same adversarial examples $x^{*}_{S}$ generated from the student network as illustrated in \cref{fig:overview}(b).
The main reason \thiswork uses the adversarial examples $x^{*}_{S}$ generated from a student network is to make peer network robust to the adversarial examples of the student network.
Though many previous works focused on the transferability of the adversarial examples among neural networks, we found that it is not necessarily true that adversarial examples have large similarities and small distances.
Therefore, we can build the peer network which is specialized in being robust to the adversarial examples of the student network while not being robust to the adversarial examples of itself.
We further describe this in \cref{sec:exp_outer_min}.
Note that the ultimate goal of our novel outer minimization process is to build a robust student network and not a robust peer network. 
Surprisingly, we find that the peer network is not robust at all to the adversarial examples aimed at itself while being robust to the adversarial examples of the student network.

\textbf{Loss function.}
To provide the soft label in inner maximization, we use the prediction of the peer model on the natural samples.
KL divergence loss is used to find the adversarial examples maximizing the discrepancy between the prediction of the peer model on natural images and the prediction of the student model on the adversarial examples :
\begin{equation}\label{inner_max_eqn}
L_{max} = KL(P_{t}(x)||S_{t}(\tilde{x}))
\end{equation} 
where $P(\cdot)$ and $S(\cdot)$ denote the prediction output of a peer model and a student model, respectively.
We use the subscript $t$ to denote the training iteration and highlight the peer network is not fixed in the process of adversarial distillation compared to other baselines.
$\tilde{x}$ is the adversarial examples which should satisfy the constraint on the magnitude of the perturbation as described in \cref{at_eqn}.

In outer minimization, the loss of the peer model consists of Cross Entropy (CE) loss and KL divergence loss.
\begin{equation}\label{peer_loss}
L_{peer} = \gamma_{1}  *H(y, P_{t}(x^{*}_{S})) \\
+ \gamma_{2}  *\tau^{2}*KL(S_{t}^{\tau}(x^{*}_{S})||P_{t}^{\tau}(x^{*}_{S}))
\end{equation}
where $\tau$ is a temperature parameter that smooths the output of a softmax layer.
The key aspect is that the peer is trained using samples adversarial to the student ($x^{*}_S$).
The cross-entropy (CE) loss  $H$ is intended to make the peer model learn the label distribution of adversarial examples $x^{*}_{S}$ with the hard label, which provides the consistent guidance.
The KL divergence loss of a peer model is to distill the knowledge of a student model, which provides the learned probability distribution of non-target classes by the student model.

The loss of the student model also consists of the cross entropy loss and KL divergence loss.
However, it also has an additional regularization term which prevents the large discrepancy between the prediction on natural images and adversarial examples \cite{trades}.
The soft label provided by the peer model is treated as the constant soft target in the loss of the student model.

\begin{equation}\label{student_loss}
\begin{split}
L_{student} &= \lambda_{1}  * H(y, S_{t}(x^{*}_{s})) \\
&+ \lambda_{2} *\tau^{2} * KL(P_{t}^{\tau}(x^{*}_{s})||S_{t}^{\tau}(x^{*}_{s})) \\
&+ \lambda_{3} *\tau^{2} * KL(S_{t}^{\tau}(x)||S_{t}^{\tau}(x^{*}_{s}))
\end{split}
\end{equation}

Then, $L_{min}$ is the sum of $L_{peer}$ and $L_{student}$.
The parameters of the two models are optimized simultaneously.
\begin{equation}
L_{min} = L_{peer} + L_{student }
\end{equation}

\section{Experimental Results}

\begin{table*}[ht]
    \centering
    \renewcommand{\arraystretch}{0.8}
    \resizebox{.7\textwidth}{!}{
    \begin{tabular}{llcccccccc}
    \toprule
    \multirow{2}{*}{Dataset} & \multirow{2}{*}{Method} & \multicolumn{4}{c}{ResNet-18}& \multicolumn{4}{c}{WideResNet34-10}\\
    \cmidrule(lr){3-6}
    \cmidrule(lr){7-10}
     & & Clean & FGSM & PGD-20 & AA & Clean & FGSM & PGD-20 & AA  \\
    \toprule
    \multirow{9}{*}{CIFAR-10} & Natural & \bf{95.42} & 35.74 & 0.00  & 0.00 & \bf{96.08} & 45.92 & 0.00  & 0.00\\
     & PGD-AT & 84.21 & 56.93 & 49.71  & 46.79 & 86.27 & 57.69 & 49.94 &  48.07\\
     & TRADES & 81.47 & 57.75 & 52.92 & 49.35 & 84.48 & 60.07 & 54.33 & 51.88\\
     & AKD$^{2}$ & 83.99 & 59.52 & 53.72 & 50.12 & 87.83 & \bf{64.14} & 56.68 & 54.25\\
     & RSLAD & 81.00 & 58.65 & \bf{54.40}  & 51.03 & 83.80 & 61.48 & 55.25 & 52.37\\
     & IAD & 80.63 & 58.13 & 53.43  & 49.88 & 83.51 & 60.91 & 54.33 & 51.89\\
     & CAT & 82.40 & 58.56 & 53.39  & 50.06 & 86.40 & 63.66 & 56.77 & 54.17\\
     & AdaAD & 81.41 & 57.45 & 53.51  & 50.08 & 84.49 & 60.65 & 55.98 & 53.38\\
     & \bf{\thiswork} & 85.01 & \bf{61.28} & 54.36  & \bf{52.57} & 85.64 & 63.40 & \bf{56.81} &\bf{55.21} \\
    \midrule
    \multirow{9}{*}{CIFAR-100} & Natural & \bf{75.48} & 8.70 &0.00   &0.00  & \bf{79.68} & 12.58 & 0.04 & 0.00 \\
     & PGD-AT & 57.30 & 28.47 & 24.15 & 21.84 & 59.57 &30.17  & 25.73 & 23.99\\
     & TRADES & 54.90 & 30.93 & 28.29  & 23.69 & 55.70 & 32.53 & 30.02 & 26.07\\
     & AKD$^{2}$ & 58.84 & 33.07 & 30.33  & 25.83 & 61.83 & 36.40 & \bf{33.20} & 28.88\\
     & RSLAD & 55.45 & 33.11 & 30.78  & 25.96 & 57.42  & 33.95 & 30.75 & 27.20\\
     & IAD & 54.98 & 32.87 & 30.28  & 25.44 &57.92 & 34.30& 31.47& 27.55\\
     & CAT & 57.81 & 33.94 & \bf{31.44}  & 25.93 & 61.68 & 36.82 & 32.95 & 28.39\\
     & AdaAD & 56.08 & 31.79 & 29.76  & 25.03 & 57.99 & 34.03 & 31.89 & 27.88\\
     & \bf{\thiswork} & 59.35 & \bf{34.41} & 29.69 & \bf{27.33} & 61.33 & \bf{37.08} & 32.39 & \bf{30.06}\\
    \midrule
    \multirow{9}{*}{TinyImageNet} & Natural & \bf{64.74} & 1.65 & 0.02 & 0.00 & \bf{68.81} & 2.32 & 0.02 & 0.00\\
     & PGD-AT & 46.25 & 24.47 & 22.53 & 17.80 & 51.10  & 27.50 & 24.83 & 20.57 \\
     & TRADES & 48.87 & 24.64 & 22.31 & 16.90 & 52.49 &27.61  & 25.36 & 19.67\\
     & AKD$^{2}$ & 50.47 & 27.25 & 25.12 & 20.01 & 54.82 & 31.83 & 29.33 & 24.09\\
     & RSLAD & 43.19 & 24.61 & 22.92 & 17.17 & 51.06 &30.28  &28.35  & 22.80\\
     & IAD &47.67& 26.11 & 23.86 &18.88 &45.96 & 27.15& 25.72& 20.80\\
     & CAT & 40.66 & 23.19 & 22.06 & 15.19 & 40.85 & 24.78 & 23.20 & 16.76\\
     & AdaAD & 49.97 & 25.79 & 23.98 &18.16  & 52.22 & 28.32 & 26.53 & 21.13\\
     & \bf{\thiswork} & 55.19 & \bf{29.42} & \bf{26.10} & \bf{21.67} &58.07  & \bf{33.04} & \bf{29.51} & \textbf{24.82}\\
    \bottomrule
    \end{tabular}
    } 
    \caption{The white-box robustness under various attack methods.
    \vspace{-10pt}} \label{tab:main_acc}
\end{table*}

\subsection{Experiment Settings}\label{exp_setting}
\textbf{CIFAR-10 and CIFAR-100~\cite{cifar10}.}
For all the baselines~\cite{pgd,trades,akd,rslad,iad, cat,adaad} results, we trained the baselines following their original settings.
For \thiswork results, we followed the training setting of \cite{rslad}.
In detail, we trained \thiswork for 300 epochs and the training batch size is 128.
The learning rate is $1e-1$ and it decays at epochs of 215, 260, and 285 by a factor of 10.
The weight decay is $2e-4$.
We applied weight averaging to \thiswork and AKD$^{2}$ following \cite{akd} for a fair comparison.
The detailed hyperparameters can be found in the supplementary materials.

\textbf{TinyImageNet~\cite{tinyimagenet}.}
We follow the hyperparameters of \cite{akd, iad} for the baselines.
For \thiswork, we trained ResNet-18~\cite{resnet} with 200 epochs and WideResNet34-10~\cite{wrn} with 100 epochs. 
The total batch size is 128.
We use the SGD optimizer with 0.9 momentum and the weight decay $2e-4$.
The initial learning rate is set to $1e-1$ and decays at epoch 100 and 150 for ResNet-18, and at epoch 50 and 80 for WideResNet34-10 by a factor of 10.
We applied weight averaging to \thiswork and AKD$^{2}$ following \cite{akd} for a fair comparison.
For more details, please refer to the supplementary materials.

\textbf{Evaluation metrics.}
We tested the white-box robustness of the baselines and \thiswork.
We report the best robust accuracy validated by PGD-10~\cite{pgd} with a step size of $2/255$ and a perturbation budget of $\epsilon = 8/255$.
Only the non-robust model, which is denoted by \textit{Natural} was chosen by the best natural accuracy because it is not robust at all in the course of training.
PGD-20 attack was conducted with a step size of $2/255$ and a perturbation budget of $\epsilon = 8/255$.
FGSM~\cite{fgsm} attack also used the same perturbation budget as PGD.
However, these attacks are not perfect for checking the robustness of a model because it is vulnerable to gradient obfuscation \cite{obfuscation}.
AutoAttack (AA)~\cite{autoattack} is prevalently regarded as the strongest attack. 
It includes targeted, untargeted PGD attacks and black-box score-based attacks~\cite{square}.

\textbf{Teacher models of baselines.}
In adversarial distillation, the performance of the student model also depends on the teacher model.
The larger models usually show better robustness than the smaller ones~\cite{pgd}, so we used the same or larger teacher network than a student in the evaluation.
Without mention, the same architecture of the teacher model is used for AD as a default. 
We adversarially trained the robust teacher model using TRADES~\cite{trades} as the teacher model because it shows better robustness than PGD Adversarial Training (PGD-AT)~\cite{pgd}.
We mainly evaluated with ResNet-18~\cite{resnet} and WideResNet34-10~\cite{wrn}, which are representative models in the adversarial robustness community.

\subsection{Adversarial Robustness Result}
\label{main_exp}
\setlength{\parskip}{0pt}
\cref{tab:main_acc} reports the white-box robustness of various baselines and \thiswork.
AutoAttack~\cite{autoattack} is the most reliable metric because many gradient-based attacks (FGSM and PGD attacks) are vulnerable to gradient obfuscation and give a false sense of security~\cite{obfuscation}.
\Thiswork shows higher AutoAttack accuracy than all the baselines from 0.73$\%p$ to 1.66$\%p$, and surpasses the clean accuracy of the other adversarial distillation baselines by up to 4.72$\%p$.
The improvement in \thiswork is more significant with ResNet-18 compared to WideResNet34-10.
This is a favorable result in adversarial distillation settings because the distillation is often conducted to improve the robustness of a small model. 
Overall, \thiswork provides a much better trade-off between the clean accuracy and the robust accuracy than the other adversarial distillation baselines.
The effect of \thiswork is not limited to small-scale datasets and small models because \thiswork also improves the result of large-scale dataset TinyImagenet with WideResNet34-10.
In \cref{sec:eval:ablation_swa}, we conduct an ablation on weight averaging which is applied to AKD$^{2}$ and \thiswork by applying it to the other baselines. 
We include the results of the transfer-based attack and gradient obfuscation tests in \cref{obfuscation_test} to exclude the possibility of gradient obfuscation.

\begin{figure*}[t]
     \centering
     \begin{minipage}{0.92\linewidth}
         \hspace{1cm}
         \includegraphics[width=0.9\linewidth]{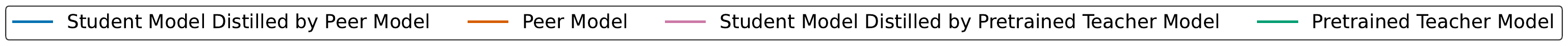}

\begin{subfigure}{.49\linewidth}
    \includegraphics[width=\linewidth]{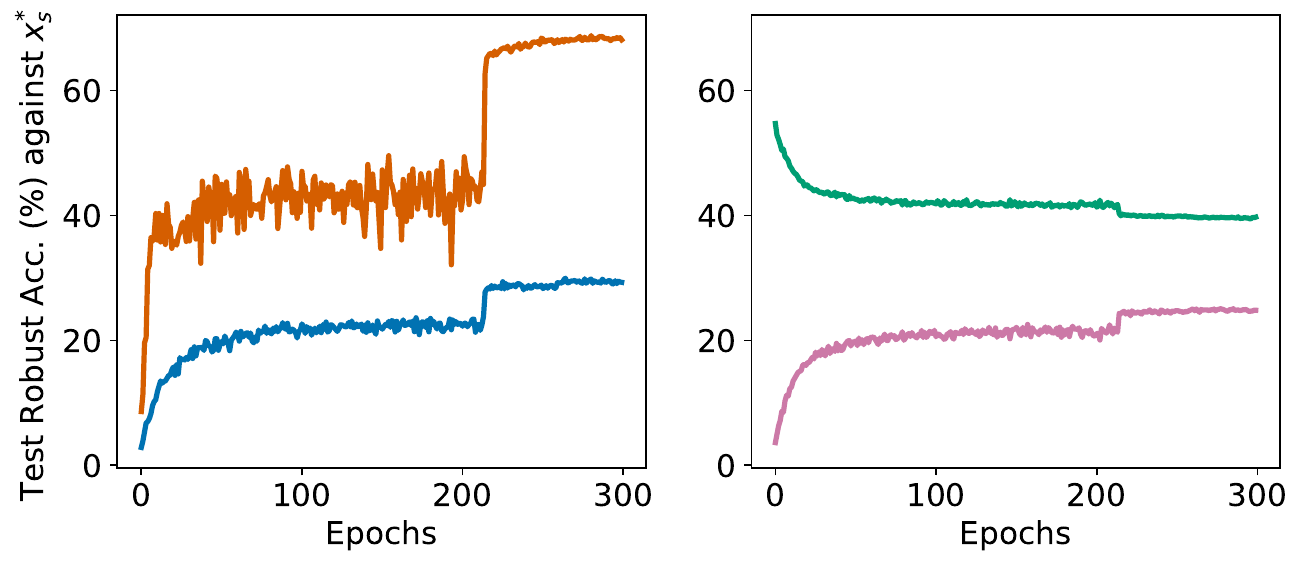}

\caption{Test robust accuracy with student-generated adversarial examples.} 
\end{subfigure}
\hfill
\begin{subfigure}{.49\linewidth}
    \includegraphics[width=\linewidth]{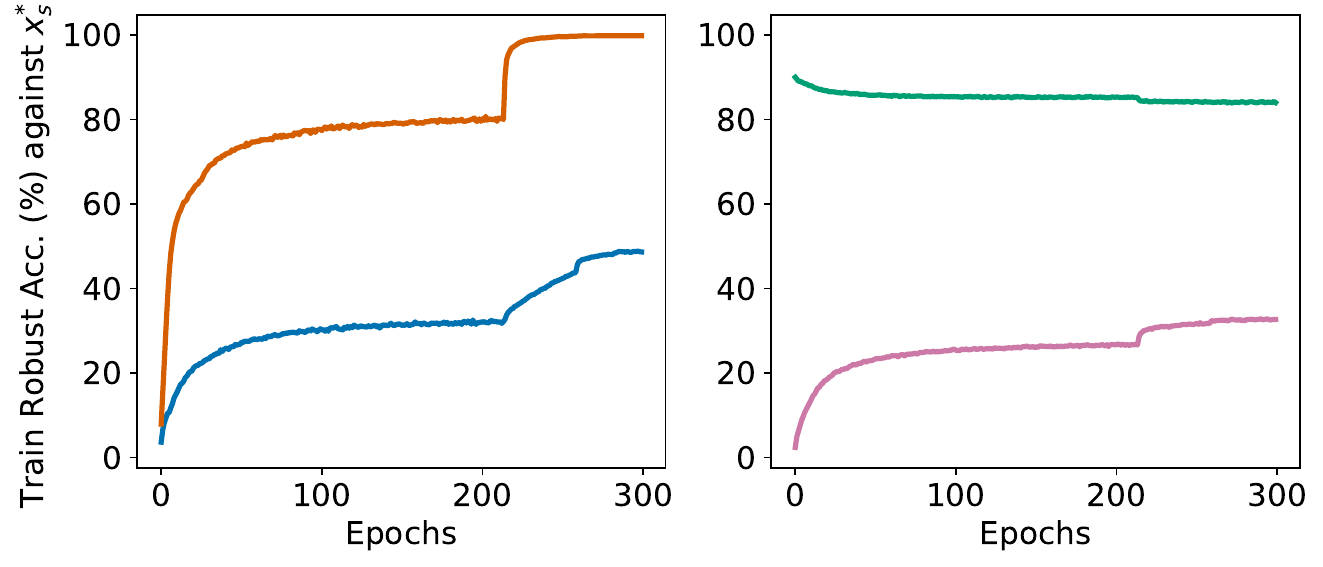}
\caption{Train robust accuracy with student-generated adversarial examples.} 
    
\end{subfigure}
\end{minipage}

\caption{Robust accuracy against student-generated adversarial examples $x^{*}_{s}$. Test (a) and train (b) robust accuracy are presented. ResNet-18 is used to measure the robust accuracy with CIFAR-100.}

\label{fig:acc_curves}
\end{figure*}

\subsection{Effectiveness of \SCHEME}
\label{sec:exp_inner_max}

\begin{table*}[ht]
    \centering
    \renewcommand{\arraystretch}{0.9}
    \resizebox{0.9\textwidth}{!}{
    \begin{tabular}{cccccccc}
    
    \toprule
     $f$ & $Clean_{f}(x)$ &  $Rob_{f}(x^{*}_{f})$ &  $Rob_{f}(x^{*}_{S})$  &  $Rob_{S}(x^{*}_{S})$ & $cos(f(x), f(x^{*}_{S}))$ & $cos(f(x^{*}_{S}), f(x^{*}_{f}))$  & $cos(f(x), f(x^{*}_{f}))$\\
    \toprule
     Peer tutor (ours) &  \textbf{75.63}   & 0.00 & \textbf{69.19} & \textbf{29.69} & 0.95 & 0.37 & 0.35\\
     Pretrained robust teacher & 57.30 & 24.15 & 39.46  & 24.48 & 0.96 & 0.97 & 0.96   \\
     Naturally pretrained teacher &  \textbf{75.48} & 0.00 & 43.05 & 14.39& 0.80 & 0.47 & 0.41  \\
    \bottomrule
    
    \end{tabular}
    }
    \caption{The relationship between the cosine similarities of the penultimate layer representation measured with ResNet-18 on CIFAR-100. $Rob_{f}(\cdot)$ denotes the robust accuracy of $f$. $S$ denotes the student network. $Clean_{f}(x)$ means the clean accuracy of $f$ on natural samples.
    \vspace{-5pt}}
    
    \label{tab:cosine}
\end{table*}

  In \cref{fig:acc_curves}, we plotted the training and test robust accuracy curves of each model 
against the adversarial examples $x^{*}_{s}$ generated from a student model to show the effectiveness of \scheme. 
The pretrained teacher model is adversarially trained beforehand with PGD-10.
The peer model is simultaneously trained with the student model with the loss of \cref{peer_loss}.
For an ablation study, we applied the same training loss to the student model with \cref{student_loss} regardless of whether it is distilled by the pretrained robust teacher or the peer model.

There are distinct patterns with peer teaching compared to the training with the pretrained teacher model.
In \cref{fig:acc_curves}(a), the test robust accuracy of the peer network against $x^{*}_{S}$ (orange) jumps up at the early epoch of adversarial distillation.
Then it experiences an additional jump right after the learning rate decay at epoch 215.
However, the test robust accuracy of the pretrained model against $x^{*}_{S}$ (green) keeps decreasing after the initial epoch of the training.
This degradation in the robust accuracy of the pretrained teacher model is due to the increasing complexity of the adversarial examples generated from a student model as the student model becomes robust along the adversarial training~\cite{iad}. 

The improvement that comes from peer teaching can be explained by the empirical robustness that the peer model attained during the adversarial distillation.
In \cref{fig:acc_curves}(b), the train robust accuracy of the peer model with the adversarial examples $x^{*}_{S}$ (orange) increases because the peer model learns $x^{*}_{S}$ directly and the weak part of the student model.
After the first learning decay at epoch 215, the train robust accuracy of the student keeps increasing. 
However, the pretrained teacher suffers from the distribution shift between the adversarial examples it learned during adversarial training and the transferred adversarial examples $x^{*}_{S}$ in the process of the adversarial distillation.
Therefore, the training robust accuracy of the pretrained teacher model on $x^{*}_{S}$ (green) keeps decreasing as illustrated in \cref{fig:acc_curves}(b).

\subsection{Robustness of Peer Network}
\label{sec:exp_outer_min}

In \thiswork, the peer model plays the role of guiding student models in both the inner maximization and outer minimization process.
As shown in \cref{tab:cosine}, we found that this peer model is specialized in defending the attack samples $x^{*}_{S}$ generated from the student model.
This peer model has higher robust accuracy against $x^{*}_{S}$ ($Rob_f(x^*_S)$) than the pre-trained robust model.
However, notably, this peer network is not robust at all, and its robust accuracy against itself with PGD-20 ($Rob_f(x^*_f)$) is $0\%$ as illustrated in \cref{tab:cosine}.

In \cref{tab:cosine}, we also measured the cosine similarity between adversarial examples and natural examples in the feature space. We intend to find how the peer model can provide reliable guidance to the student model while it is not robust at all against the adversarial examples aimed at itself. 
The peer model shows a comparable cosine similarity between natural samples $x$ and $x^{*}_{S}$ in the features space 
compared to the pre-trained robust teacher.
It embeds $x^{*}_{S}$ around $x$ and this is the desirable property of the robust model because the robust model is expected to make the prediction on the adversarial examples equal to the natural samples.
\setlength{\parskip}{0pt}

However, the peer model has much higher natural accuracy which even slightly surpasses the naturally pretrained teacher.
The peer model achieves a natural accuracy of $75.63\%$, significantly higher than the $57.30\%$ achieved by the pretrained teacher model.
This superior performance the peer model shows with natural examples and $x^{*}_{S}$ implies that it is a better approximator for the label distribution of natural samples and $x^{*}_{S}$ than the pretrained robust teacher model. 
The pretrained robust teacher inevitably suffers from the degradation in the natural accuracy compared to the standard training due to the trade-off between the clean accuracy and the robust accuracy~\cite{trades, tradeoff_odds, reconcile}, whereas the peer model does not experience this trade-off.

\subsection{Ablation on Weight Averaging}
\label{sec:eval:ablation_swa}

\begin{table}
    \centering
    \renewcommand{\arraystretch}{0.85}
    \resizebox{.76\columnwidth}{!}{
    \begin{tabular}{lcccc}
    \toprule
      \makecell[c]{Method}  & Clean & FGSM & PGD-20 & AA\\
     \midrule
       Natural$+$SWA& \textbf{67.42} & 2.22 & 0.04 & 0.00 \\
       PGD-AT$+$SWA& 49.10 & 26.32 & 23.95 & 19.46 \\
       TRADES$+$SWA& 50.22 & 26.10 & 23.79 & 18.47 \\
       RSLAD$+$SWA& 42.13 & 24.33 & 22.99 & 17.32 \\
       IAD$+$SWA& 48.70 & 26.94 & 24.88 &19.62  \\
       CAT$+$SWA& 38.13 & 22.13 & 20.99 & 14.53 \\
        AdaAD$+$SWA& 50.41 & 25.81 & 24.20 & 18.43 \\

        \thiswork w/o SWA& 54.01 & 28.21 & 25.13 & 20.00 \\
        \thiswork & 55.19 & \textbf{29.42} & \textbf{26.10} & \textbf{21.67} \\

    \bottomrule
    \end{tabular}
    } 
    \caption{Ablation of SWA with ResNet-18 on TinyImageNet
    \vspace{-10pt}}\label{ablation_of_swa}
\end{table}

We include the ablation of Stochastic Weight Averaging (SWA) to check the effectiveness of \thiswork without SWA.
In \cref{ablation_of_swa}, 
\thiswork without SWA shows the higher clean accuracy than adversarial training baselines which incorporate SWA by up to 3.6\%p, while also exhibiting superior robust accuracy with ResNet-18 on TinyImageNet.
We applied weight averaging to the baselines following \cite{akd}, which applies SWA from the epoch of the first learning rate decay to the last epoch.
SWA also increases the robust accuracy of the adversarial training baselines except for CAT.
The results in \cref{ablation_of_swa} indicate that SWA is not an essential part of \thiswork but complements it.

\subsection{Teacher Sensitivity}
\label{sec:eval:teacher_sensi}

\begin{table}
    \centering
    \renewcommand{\arraystretch}{0.85}
    \resizebox{.75\columnwidth}{!}{
    \begin{tabular}{lcccc}
    \toprule

     Method & Clean & FGSM  & PGD$_{20}$ & AA  \\
    \toprule
     AKD$^2$ & 57.33 &32.89 & 30.26 &25.69\\
     RSLAD &55.39& 32.21& 29.29 &24.49\\
     IAD &55.51& 31.44& 28.54 &24.41\\
     CAT &56.61 &34.06 & 31.16 & 25.50\\
     AdaAD & 56.72& 31.95& 29.01 & 24.86\\
     \bf{\thiswork} &57.63&34.33& 30.17 & \textbf{26.99}\\

    \bottomrule
    \end{tabular}
    } 
 \caption{CIFAR-100 robust accuracy of ResNet-18 with WRN34-10 teacher (peer) model.
 \vspace{-5pt}}\label{tab:large_teacher_acc}    
\end{table}

\begin{table}
    \centering
    \renewcommand{\arraystretch}{0.85}
    \resizebox{.75\columnwidth}{!}{
    \begin{tabular}{lccccc}
    \toprule

     Method & Clean & FGSM  & PGD$_{20}$ & AA  \\
    \toprule
     AKD$^2$ &82.75 & 57.03& 52.68  & 48.45\\
     RSLAD & 79.91 & 57.25& 53.54  &49.85\\
     IAD &80.15& 57.97 & 53.23 & 49.10\\
     CAT &77.22 &53.13 & 49.37  &45.17\\
     AdaAD & 79.81 & 54.81& 51.21  &47.59\\
     \bf{\thiswork} & 82.41& 57.43& 52.00 & \textbf{50.02}\\
    \bottomrule
    \end{tabular}
    } 
 \caption{CIFAR-10 Robust accuracy of MobileNetV2.
 \vspace{-10pt}}\label{mobilenet_table}
\end{table}

 Some works~\cite{akd, cat, iad} assume a situation where a teacher model and a student model have the same capacity. 
However, other approaches~\cite{rslad, adaad} chose a larger teacher model than a student model to measure its effectiveness.
For a fair comparison, we also tested the effectiveness of \thiswork with the large teacher (peer) model on CIFAR-100 dataset.
In ~\cref{tab:large_teacher_acc}, the teacher (peer) model is WideResNet34-10, and the student model is ResNet-18.

We observed that the student model trained with \thiswork still maintains improved performance compared to the baselines with a large teacher model. 
These results support the proposed method of \thiswork is not limited to the case when the peer network has to be the same architecture.
In \cref{mobilenet_table}, we also tested the effectiveness of \thiswork with MobileNetV2 \cite{mobilenetv2} on CIFAR-10, a widely used neural network in distillation settings.
\Thiswork also shows the highest AutoAttack accuracy and better tradeoff between the robustness and the clean accuracy with MobileNetV2.

\subsection{Gradient Obfuscation Test}\label{obfuscation_test}

It has been highlighted that any robust model should be secure against transfer-based attacks~\cite{cw, obfuscation, blackbox}.
Therefore, it must be checked whether \thiswork shows robustness against transfer-based attacks.
Here, we train two surrogate models, which are ResNet-34~\cite{resnet} and MobileNetV2~\cite{mobilenetv2} with PGD-10 adversarial training~\cite{pgd}.
The training setting is the same as the PGD-AT baseline in \cref{exp_setting}.
We transferred the adversarial examples generated from these two surrogate models to ResNet-18 which is trained by each method.
FGSM and PGD-20 were used to create adversarial examples and the robust accuracy of models trained by each method are described in \cref{tab:transfer_acc} 
It shows that \thiswork is more robust than baselines against transfer-based attacks.

It is also known that previous works on adversarial training actually rely on the obfuscated gradient, giving a false sense of security~\cite{obfuscation}.
We conducted the gradient obfuscation test mentioned in~\cite{obfuscation}.
First, the robust accuracy of \thiswork against PGD-10 is similar to the robust accuracy against PGD-1K in \cref{tab:gradient_obfuscation}.
Second, the unbounded attack on \thiswork successfully reaches $0\%$ robust accuracy in \cref{tab:gradient_obfuscation}.
Third, \cref{tab:main_acc} shows that the success rate of a one-step attack is lower than PGD-20 with \thiswork.
It implies that the inner maximization process is not stuck in a local solution. 
Lastly, AutoAttack~\cite{autoattack} includes SQUARE attack ~\cite{square}, a black-box score-based attack.
Therefore, the above results exclude the possibility of gradient obfuscation in \thiswork.

\begin{table}
    \centering
    \renewcommand{\arraystretch}{0.85}
    \resizebox{.8\columnwidth}{!}{
    \begin{tabular}{lcccc}
    \toprule
    Surrogate & \multicolumn{2}{c}{ResNet-34} & \multicolumn{2}{c}{MobileNetV2}\\
    \cmidrule(lr){1-5}
     Method & FGSM & PGD$_{20}$  & FGSM & PGD$_{20}$  \\
    \toprule
     PGD-AT & 38.54 & 37.11& 38.62 & 37.08\\
     TRADES & 38.84 & 38.02& 38.16 & 37.34\\
     AKD$^2$ &40.95& 39.77& 38.89 & 37.88\\
     RSLAD & 40.12& 39.29& 38.85 &37.82 \\
     IAD &39.63 & 38.81 & 39.16 & 37.97\\
     CAT &42.38 & 41.66& 39.46& 38.42\\
     AdaAD & 39.24 &38.30 & 38.60 & 37.27\\
     \textbf{\thiswork} & \textbf{44.23} & \textbf{43.61}& \textbf{42.15} &\textbf{40.73} \\
    \bottomrule
    \end{tabular}
    } 
 \caption{Checking gradient obfuscation by measuring the robust accuracy of ResNet-18 on CIFAR-100 dataset under transfer-based attacks.
 \vspace{-5pt}}\label{tab:transfer_acc}    
\end{table}

\begin{table}
    \centering
    \renewcommand{\arraystretch}{0.85}
\resizebox{.85\columnwidth}{!}{
    
    \begin{tabular}{ccccccccc}
    \toprule
    Dataset & Model & PGD-10 & PGD-1K & $\epsilon=\infty$\\
     \toprule
     \multirow{2}{*}{CIFAR-10} & ResNet-18 &55.54 &53.94  &0.00 \\
     & WRN34-10 &57.89 & 56.43 &0.00 \\
     \midrule
     \multirow{2}{*}{CIFAR-100} & ResNet-18 &30.66&29.36&0.00 \\
     & WRN34-10 &33.24 & 31.99 & 0.00\\
     \midrule
     \multirow{2}{*}{TinyImageNet} & ResNet-18 & 26.34 & 25.86 &0.00\\
     & WRN34-10 & 29.99 & 29.34 & 0.00 \\

    \bottomrule
    \end{tabular}
    }
 \caption{Obfuscated gradient test results proposed in \cite{obfuscation}. \vspace{-10pt}}\label{tab:gradient_obfuscation}    
\end{table}

\subsection{Loss Landscape Visualization}

\begin{figure}
     \centering
     \begin{subfigure}[b]{.265\linewidth}
         \centering
         \includegraphics[width=\columnwidth]{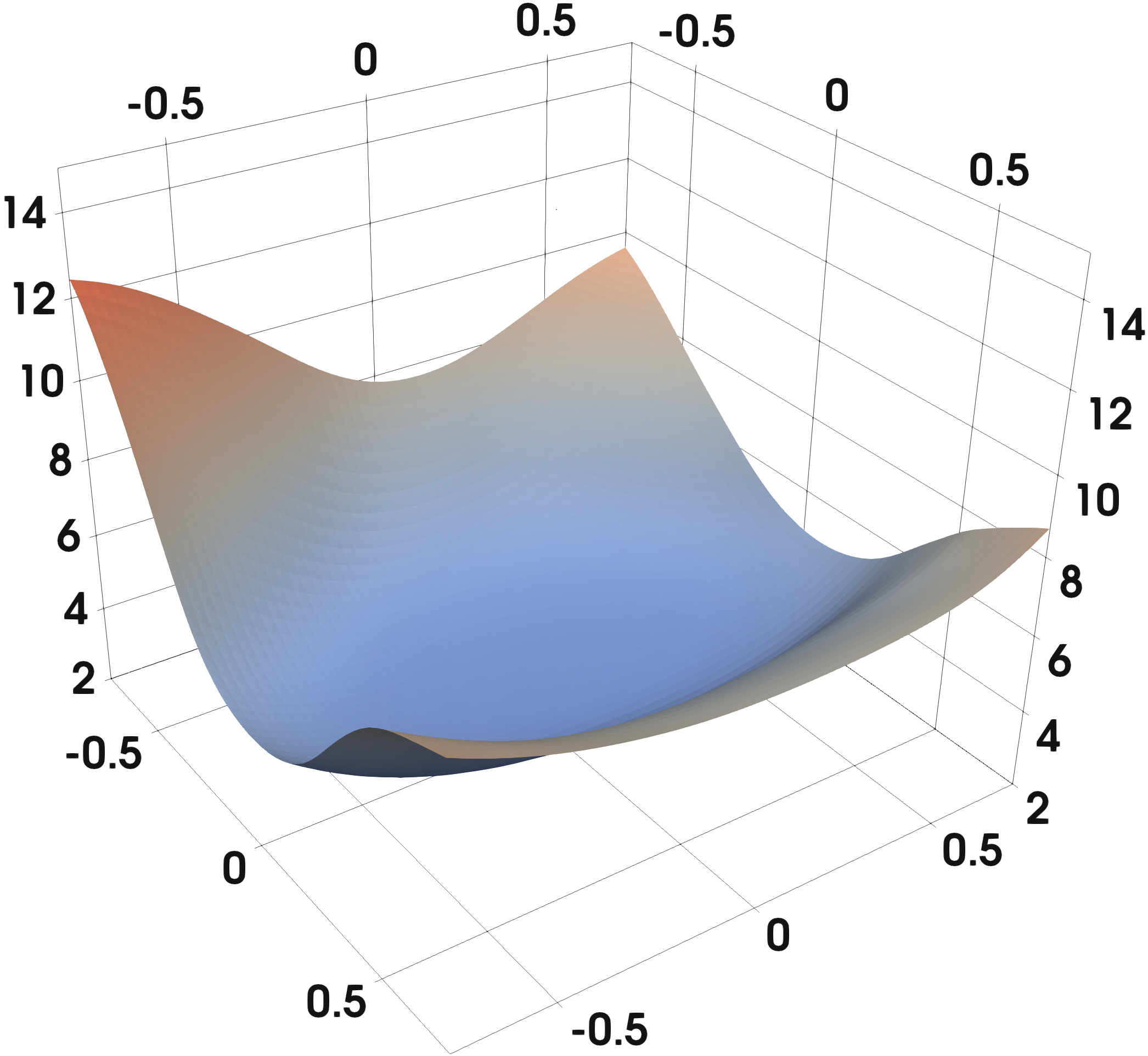}
         \caption{PGD-AT}
         \label{fig:lla}
     \end{subfigure}
     \hfill
     \begin{subfigure}[b]{.265\linewidth}
         \centering
         \includegraphics[width=\columnwidth]{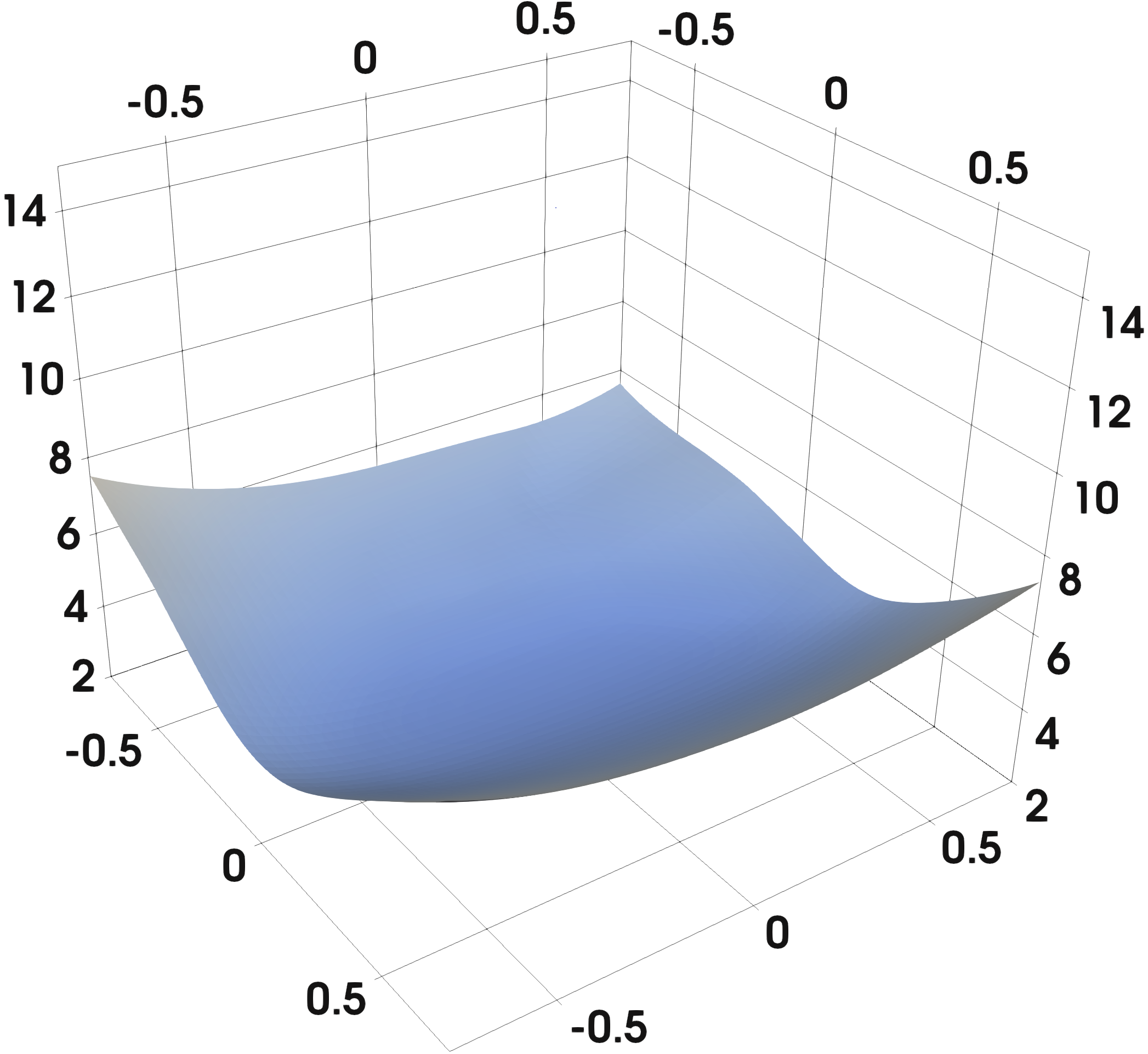}
         \caption{TRADES}
         \label{fig:llb}
     \end{subfigure}
     \hfill
     \begin{subfigure}[b]{.265\linewidth}
         \centering
         \includegraphics[width=\columnwidth]{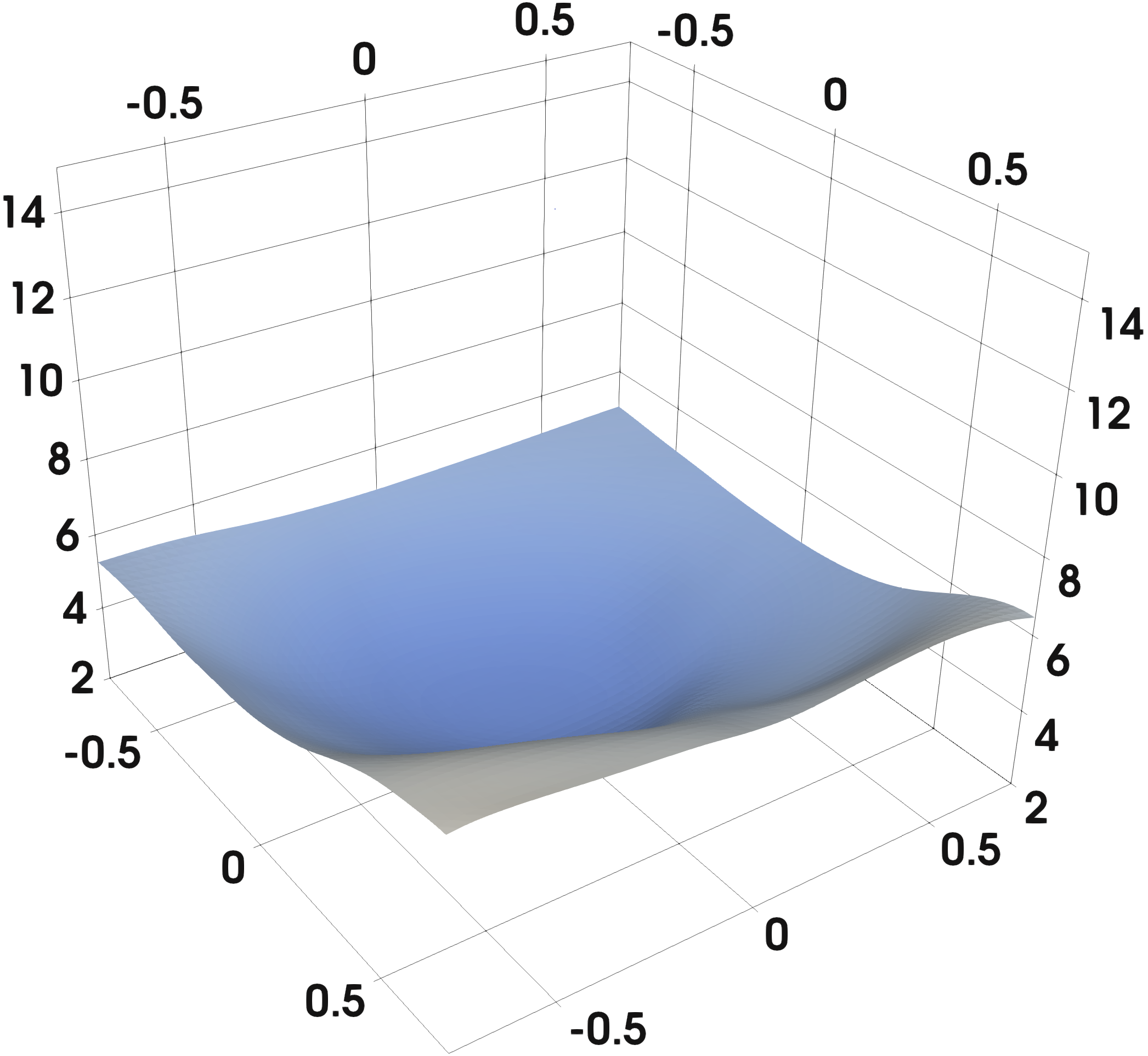}
         \caption{\thiswork (Ours) }
         \label{fig:llc}
     \end{subfigure}
     \hfill
     \begin{subfigure}[b]{.07\linewidth}
         \centering
         \includegraphics[width=\columnwidth]{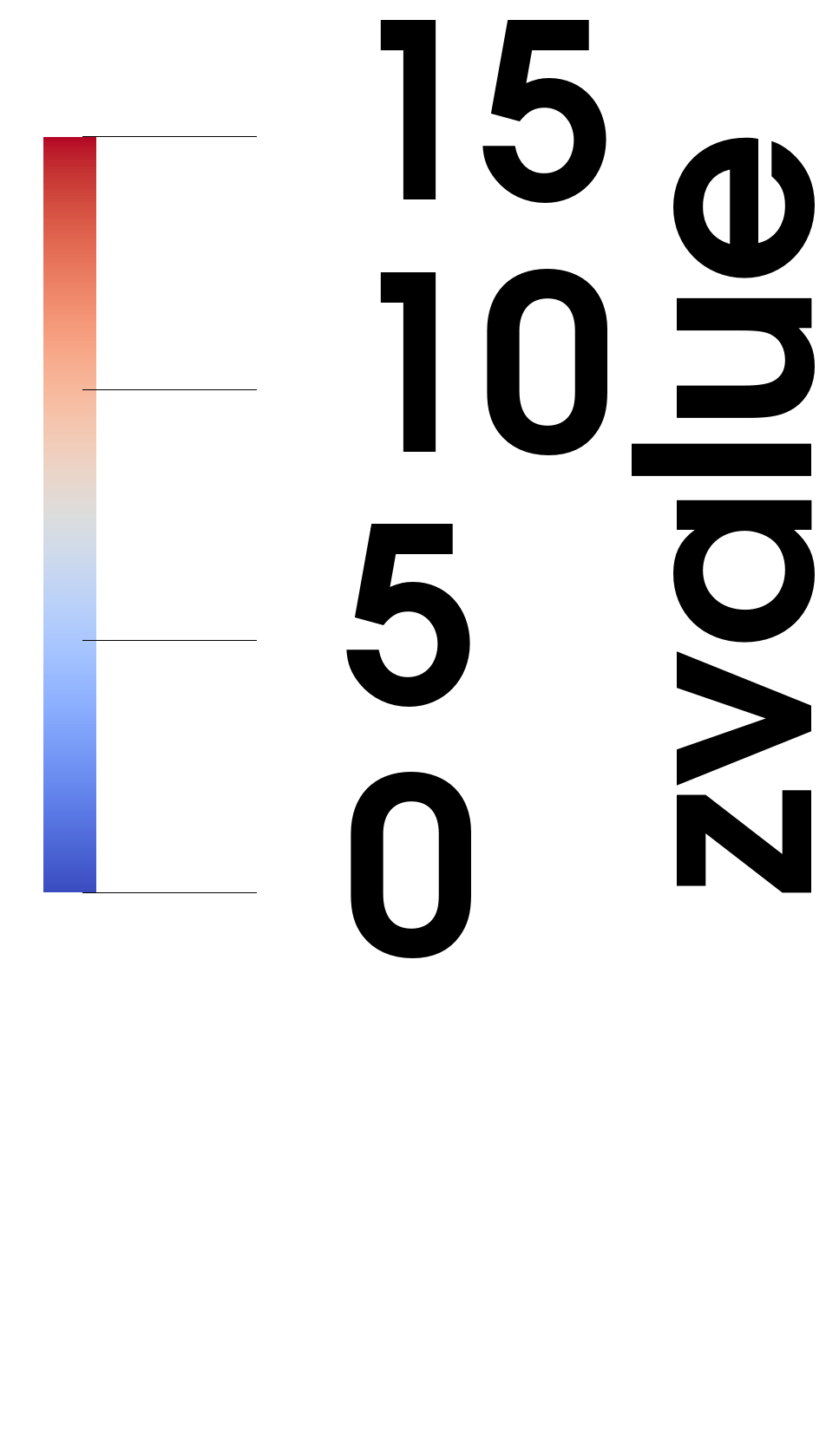}        
     \end{subfigure}
        \vspace{-5pt}
        \caption{Comparison of weight loss landscape visualization~\cite{visualloss} between baselines~\cite{pgd, trades} and \thiswork. The WRN34-10~\cite{wrn} model trained with CIFAR-100 by each method is perturbed along a random direction within the range of [$-0.75$, $0.75$]. The vertical axis $z$ denotes the loss value.
        \vspace{-10pt}
        }
        \label{fig:three graphs}
\end{figure}

Prior works~\cite{lossland3} found that a flat loss landscape is favorable to the generalization of neural networks. 
\cite{sam_min} showed that a sharp loss landscape causes a big difference between training and test distribution. 
Especially in the context of adversarial robustness, previous works~\cite{akd} showed that a flatter loss landscape helps to mitigate robust overfitting. 
In this regard, many previous works about adversarial robustness~\cite{adaad, akd, lossland2, lossland3, lossland4} showed that their method makes the loss landscape flatter as expected. 
The loss landscape of \thiswork also coincides with these arguments. 
In \cref{fig:three graphs}, we visualize the loss landscape in weight space. 
Compared to \thiswork, the loss landscape of PGD-AT and TRADES is sharper than the one of \thiswork. 
The flat loss landscape of \thiswork explains the generalization ability and the adversarial robustness of the student network against the adversarial examples generated from the unseen test dataset.

\subsection{Visualization of Feature Representation}

\begin{figure}
     \centering
         \includegraphics[width=\columnwidth]{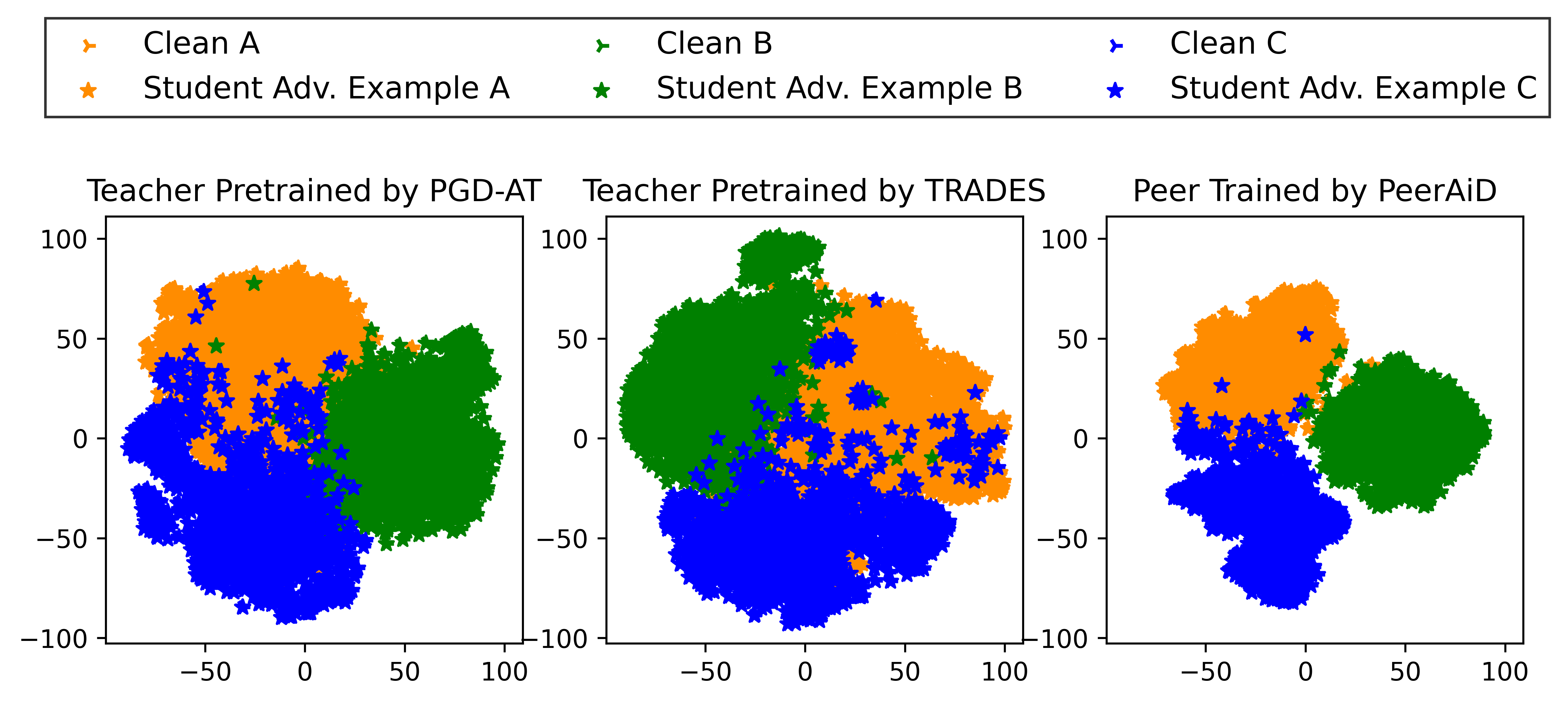}
         \vspace{-15pt}
         \caption{t-SNE results of the penultimate layer representation with the pretrained robust teacher model and \thiswork. 
         \vspace{-15pt}}
         \label{fig:t_sne_result}
\end{figure}

In \cref{fig:t_sne_result}, we visualize the feature representation of the teacher (peer) model with the adversarial examples $x^{*}_{S}$ generated from the student model.
Two ResNet-18 are adversarially pretrained with PGD-AT and TRADES, respectively.
Three classes from CIFAR-10 are randomly chosen for better visualization.
The peer model trained by \thiswork has a better ability to embed the natural examples and transferred adversarial examples $x^{*}_{S}$ because it clearly separates one class from other classes in \cref{fig:t_sne_result}, which is consistent with \cref{sec:exp_outer_min}.
On the other hand, the feature representation of the pre-trained robust teacher model shows more overlaps among classes.

\subsection{Visualization of Semantic Gradients}
\begin{figure}[t]
\centering
\includegraphics[width=.65\columnwidth]{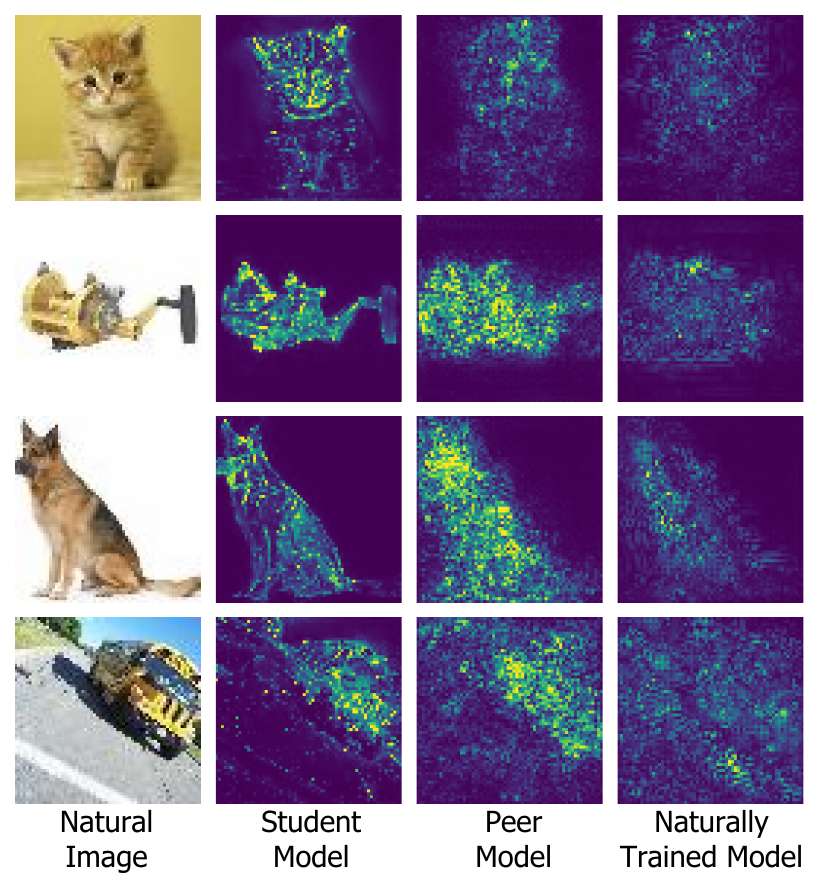}
        \vspace{-8pt}
        \caption{Visualization of semantic gradients. We attack each model and get gradients $\nabla_{x}L$. Within the same attacked model, we clip pixels to $\pm$3 standard deviations and sum the absolute pixel values across channels. We scale the values to [0,1].
        \vspace{-15pt}}\label{fig:semantic_grad}
\end{figure}

It is generally perceived that adversarially robust models show semantic or interpretable gradient with respect to inputs $\nabla_{x}L$~\cite{reconcile,free,tradeoff_odds}.
We observe the same phenomenon with the student model trained by \thiswork in \cref{fig:semantic_grad}.
We visualized the semantic gradient of three models (student, peer, and non-robust model) to find distinct patterns among models with TinyImageNet dataset and ResNet-18.

In the second column of \cref{fig:semantic_grad}, pixels along the edge of the objects in the pictures have a large magnitude of gradients with respect to input with the student model.
It indicates that the edge or unique pattern of the object needs to be attacked to fool the robust student model~\cite{shapley}.
In contrast, the last column shows that the high-magnitude gradients are spread across broad regions, showing that non-robust models can be easily fooled by attacking any pixels.
Note that the values in \cref{fig:semantic_grad} are normalized individually within each column of the same model, and their brightness cannot be directly compared between columns.
In the third column of \cref{fig:semantic_grad}, we also observe an interesting pattern of the gradient of inputs with the peer model trained with \thiswork.
Although the peer model is not robust at all against the adversarial examples aimed at itself as illustrated in \cref{tab:cosine}, the gradient of input coming from the peer model also exhibits a similar pattern to the student model.
This can be interpreted as the peer model has some knowledge of defense similar to the student model.

\section{Conclusion}
We propose a novel online adversarial distillation method \thiswork which significantly boosts the robust accuracy. 
We found that it is possible to build a \textit {peer model} not being robust at all against the white-box attack while being much more robust to the attack examples of the student network. 
The peer network is specialized in defending the attack samples of the student network and this leads to the more reliable guidance of the peer network than the pretrained robust model used in conventional methods.
With peer tutoring, we improved both the robust accuracy and natural accuracy of the student network compared to various baselines. 

\noindent\textbf{Acknowledgement.}
This work was supported by the New Faculty Startup Fund from Seoul National University (50\%) and Culture, Sports and Tourism R\&D
Program through the Korea Creative Content Agency grant funded by
the Ministry of Culture, Sports and Tourism in 2024 (CR202104003, 50\%)

\newpage
{
    \small
    \bibliographystyle{ieeenat_fullname}
    \bibliography{main}
}

\clearpage
\setcounter{page}{1}
\maketitlesupplementary

\section{Appendix}
\label{sec:appendix}

\subsection{Detailed Description of Experimental Settings}
\textbf{\thiswork.}
For CIFAR-10, we set $\gamma_{1}=1$, $\gamma_{2}=0.1$, $\lambda_{1}=1$, $\lambda_{2}=0$, $\lambda_{3}=1$.
The temperature for the peer loss and the student loss are 1 and 5, respectively.
We used the same hyperparameters for ResNet-18 and WideResNet34-10.
For CIFAR-100, we used the same hyperparameters as CIFAR-10 except for $\gamma_{2}$.
We only changed $\gamma_{2}$ to 1 because we found that the knowledge of a student model is useful for the dataset with a large number of classes. 
We also used the same hyperparamters for ResNet-18 and WideResNet34-10 with CIFAR-100.
For TinyImageNet, we set $\gamma_{1}=1$, $\gamma_{2}=100$, $\lambda_{1}=35$, $\lambda_{2}=0.035$, $\lambda_{3}=20$.
The temperatures of the peer loss and student loss  are 1 and 1, respectively.
We used the same coefficients and the temperature in the loss terms for ResNet-18 and WideResNet34-10 with TinyImageNet.
In all our experiments, the peer model and student model used the same parameters of training epoch, learning rate, batch size, and weight decay.

\textbf{Baselines.}
We followed their original settings for baselines as mentioned in \cref{exp_setting}.
In detail, there are two versions of IAD~\cite{iad} in the original paper.
IAD presented IAD-I and IAD-II depending on whether a naturally trained teacher is used or not.
We chose IAD-I in all experiments because the paper mentioned IAD-I generally shows better robustness with the teacher model pretrained by TRADES.
We also tested IAD-II with WideResNet34-10 on CIFAR-100 and found a consistent result which shows a higher robust accuracy of IAD-1 than IAD-II with the robust teacher pretrained by TRADES.
CAT~\cite{cat} mentions that any two adversarial training methods can be used to train two student models collaboratively.
We chose TRADES \cite{trades} and Adversarial Logit Paring (ALP) \cite{ALP} to train the two student models because the paper mentioned TRADES and ALP show superior performance in terms of AutoAttack (AA) robust accuracy.
We measured the robust accuracy of both student models with AutoAttack and reported the higher robust accuracy between the two student models for CAT.

\subsection{Robustness against Other Attacks}

\begin{table}
    \centering
    \resizebox{0.99\columnwidth}{!}{
    \begin{tabular}{lcccc}
    \toprule
    \multirow{2}{*}{Method} & \multicolumn{2}{c}{ResNet-18} & \multicolumn{2}{c}{WideResNet34-10}\\
    \cmidrule(lr){2-5}
      & CW$_{2}$ & MI-FGSM  & CW$_{2}$ & MI-FGSM  \\
    \toprule
     PGD-AT & 44.76 & 23.80 &49.26 &26.81 \\
     TRADES &46.55& 23.89& 50.46 & 26.91\\
     AKD$^2$ & 48.46& 26.47& 52.43 & 30.52\\
     RSLAD & 41.81& 24.04& 44.48 & 26.46\\
     IAD &45.82 & 25.15 & 49.00 & 29.32\\
     CAT &39.10 &22.58 & 39.53& 24.01\\
     AdaAD & 47.90 & 25.08& 49.96 & 27.61\\
     \textbf{\thiswork} & \textbf{53.13} & \textbf{28.03}& \textbf{55.63} &\textbf{31.32} \\
    \bottomrule
    \end{tabular}
    } 
 \caption{Test robust accuracy of the models trained by the baselines and \thiswork against CW$_{2}$ and MI-FGSM attack with TinyImageNet.\vspace{-10pt}}\label{tab:other_attack}    
\end{table}
We checked the robustness of \thiswork and the baselines with two additional attacks.
We chose CW$_{2}$ attack \cite{cw} and MI-FGSM attack \cite{mi_fgsm} to check the robustness of baselines and \thiswork.
Overall, \thiswork shows higher robust accuracy against them, as illustrated in \cref{tab:other_attack}.
In particular, \thiswork improves the robust accuracy against CW$_{2}$ attack by up to $3.2\%p$ with WideResNet34-10 on TinyImageNet.
CW attack uses margin-based loss and minimizes the norm of the perturbation.
We conducted CW$_{2}$ attack following \cite{adaad} and set the balance constant $c$ to 0.1.
We chose $l_{2}$ norm for the norm of the perturbation in CW attack because the original paper \cite{cw} mentioned the defenders should show the robustness against $l_{2}$ attack.
MI-FGSM attack is a momentum-based iterative method to find adversarial examples.
We set the iteration number to 10 and the decaying factor to 1 with MI-FGSM attack.
All other experimental settings are the same as in \cref{exp_setting}.

\subsection{Difference from the Prior Art}
Many previous works which aim at adversarial distillation require a pretrained robust model, whereas \thiswork does not require the pretrained robust model.
CAT \cite{cat} proposed online adversarial distillation which also collaboratively trains two student models.
However, \thiswork significantly differs from CAT regarding the inner maximization process.
CAT independently attacks two student models with different attack methods because their approach is based on the idea that each student model trained by distinct attack methods learns different features.
On the other hand, \thiswork attacks only a single student model, and thus, the computational cost of adversarial distillation is twice times smaller than CAT, as emphasized in \cref{tab:training_time}.
In addition, while previous works did not focus on the non-transferability of adversarial examples, \thiswork considers this aspect.
Specifically, as illustrated in \cref{tab:cosine}, the adversarial examples generated from the student model are not strong enough to fool the peer model because the peer model becomes the specialist who defends the adversarial examples aimed at the student model.  
Based on the above finding, \thiswork lets the peer model guide the student model.

\subsection{Evaluation on Different Models}

\begin{table}
    \centering
    \begin{tabular}{llcccc}
    \toprule
    \multirow{2}{*}{Method} & \multicolumn{2}{c}{DenseNet-BC-40}& \multicolumn{2}{c}{DenseNet-40}\\
    \cmidrule(lr){2-5}
       & Clean &   AA & Clean &  AA  \\
    \toprule
      Natural & \textbf{92.75} & 0.00 & \textbf{94.39} & 0.00 \\
      PGD-AT  & 77.01 & 41.84  & 80.79 & 44.82 \\
      TRADES  & 73.73 & 39.78  & 76.81 & 44.81 \\
      AKD$^{2}$ & 75.19 & 41.70  & 78.41 & 46.16 \\
      RSLAD  & 71.63 & 41.00  & 74.94 & 46.05 \\
      IAD  & 71.96 & 42.48  & 76.34 & 46.64 \\
      CAT  & 73.08 &  41.07 & 75.91 & 44.77 \\
      AdaAD  & 72.63 & 39.45  & 75.80 & 43.62 \\
      \bf{\thiswork} & 75.93 & \textbf{43.26}  & 78.21 & \textbf{47.15} \\
    
    \bottomrule
    \end{tabular}
    \caption{Test robust accuracy and clean accuracy of DenseNet-BC-40 and DenseNet-40 models trained by \thiswork and the baselines on CIFAR-10.} \label{tab:densenet_acc}
\end{table}

We also checked the effectiveness of \thiswork with additional models.
We compared \thiswork and baselines with DenseNet-BC-40 and DenseNet-40 \cite{densenet}.
The number of layers $L$ was set to 40 and the growth rate $k=12$ was chosen.
As illustrated in \cref{tab:densenet_acc}, \thiswork shows higher AutoAttack (AA) robust accuracy than the baselines with DensNet and DenseNet-BC.
DenseNet-BC has fewer feature maps than DenseNet, and it is a compressed version of DenseNet.
Therefore, the robustness of DenseNet-BC is lower than that of DenseNet because DenseNet-BC has fewer parameters and a smaller model capacity \cite{pgd}.
\thiswork shows higher robust accuracy and a better trade-off between robustness and clean accuracy with both DenseNet-BC and DenseNet.
\thiswork improves the AutoAttack robust accuracy of DenseNet-BC-40 by up to $0.78\%p$ and DenseNet-40 by up to  $0.51\%p$.

\subsection{Larger Search Radius of $\epsilon$}

\begin{table}
    \centering
    \resizebox{0.99\columnwidth}{!}{
    \begin{tabular}{lcccc}
    \toprule
    Perturbation budget & \multicolumn{4}{c}{$\epsilon=10/255$} \\
    \cmidrule(lr){1-5}
     Method & Clean & FGSM  & PGD-20 & AA  \\
    \toprule
     PGD-AT & 55.15&29.73 &26.69 &23.66 \\
     TRADES &51.82 &30.25&27.83 & 23.46\\
     AKD$^2$ &\textbf{56.65} &33.49 &30.88 &26.71\\
     RSLAD &54.53 &33.30 &31.09 &25.99\\
     IAD &53.67 &33.01 & 30.89&26.17\\
     CAT &55.19& \textbf{34.14}& \textbf{32.17}&26.50\\
     AdaAD &55.72 &31.91 &29.97 &25.29\\
     \textbf{\thiswork} & 54.39 & 33.45& 30.20 &\textbf{27.19} \\
    \bottomrule
    \end{tabular}
    } 
 \caption{Test robust accuracy of ResNet-18 trained by various methods with the large perturbation budget $\epsilon=10/255$ on CIFAR-100.}\label{tab:large_epsilon}    
\end{table}
We tested the baselines and \thiswork with the student model trained on adversarial examples generated using the larger perturbation budget of $\epsilon = 10/255$.
In \cref{exp_setting}, we compared the baselines and \thiswork with the student model trained on the adversarial examples generated using the perturbation budget of $\epsilon = 8/255$.
All other training settings are the same as in \cref{exp_setting}.
We kept the perturbation budget at $\epsilon = 8/255$ during testing.
As illustrated in \cref{tab:large_epsilon}, \thiswork shows superior AutoAttack (AA) robust accuracy with the student model trained on the adversarial examples generated using a large perturbation budget.
We conducted adversarial training and adversarial distillation with ResNet-18 on CIFAR-100.
All baselines and \thiswork harm clean accuracy with a large perturbation budget due to the trade-off between robustness and clean accuracy. 
The decrease in the clean accuracy with a large perturbation budget is the smallest in AdaAD \cite{adaad} as mentioned in the original paper.
However, \thiswork still surpasses the AA robust accuracy of AdaAD by a large margin of 1.9$\%p$.

\subsection{Mitigating Robust Overfitting}

\begin{table}
    \centering
        \resizebox{0.99\columnwidth}{!}{
    \begin{tabular}{lcccccc}
    \toprule
     \multirow{2}{*}{Method} & \multicolumn{3}{c}{AA Robust Accuracy} & \multicolumn{3}{c}{Clean Accuracy}\\
    \cmidrule(lr){2-7}
      & Best & Final  & Diff & Best & Final  & Diff  \\
    \toprule
     PGD-AT &21.84 &18.61&3.23 & 57.30& 55.92& 1.38\\
     TRADES & 23.69& 23.65& 0.04& 54.90&55.39 & $-0.49$ \\
     AKD$^2$ &25.83 &25.43 & 0.40&58.84 &59.82& $-0.98$\\
     RSLAD & 25.96&26.03 & -0.07& 55.45& 55.28& 0.17\\
     IAD &25.44 &25.22 & 0.22& 54.98 &55.42 &  $-0.44$\\
     CAT &25.93 &25.11 & 0.82&57.81 & 58.48& $-0.67$\\
     AdaAD &25.03& 24.79& 0.24& 56.08& 56.29& $-0.21$\\
     \textbf{\thiswork} & 27.33& 27.28& 0.05 & 59.35 & 59.38& $-0.03$ \\
    \bottomrule
    \end{tabular}
    } 
 \caption{Robust overfitting comparison of various methods on CIFAR-100 with ResNet-18. The best checkpoint was selected based on the test robust accuracy using PGD-10.}\label{tab:robust_overfit}   
\end{table}

It is known that robust overfitting is prevalent in adversarial training~\cite{robust_overfit, akd}, and many previous works of adversarial training can be defeated by early-stopping.
Robust overfitting is the phenomenon where the test robust accuracy peaks shortly after the first learning rate decay and then degrades until the last training epoch.
\cref{tab:robust_overfit} shows that \thiswork exhibits less robust overfitting than most of the baselines.
Robust overfitting is often measured with the difference between the test robust accuracy of the best and last checkpoints.
\thiswork shows higher robust accuracy at both the best and last checkpoint of the student model than the baselines.
AutoAttack robust accuracy of \thiswork at the last training epoch is even higher than the robust accuracy of all other baselines at their best epochs.
This superior performance of \thiswork comes from the fact that \thiswork effectively mitigates robust overfitting with only 0.05$\%p$ difference in the robust accuracy between the best checkpoint and the last checkpoint.
As illustrated in \cref{fig:acc_curves}(a), the test robust accuracy of the student model trained by \thiswork does not suffer from robust overfitting in the experiment.
The test robust accuracy of the student model distilled by the peer model does not reach its peak shortly after the first learning rate decay at epoch 215.
The best checkpoint is attained at epoch 265 with \thiswork in \cref{fig:acc_curves}(a).
The best robust and clean accuracy of RSLAD is lower than those of the last checkpoint because the best checkpoint was chosen based on PGD-10 test robust accuracy, whereas the robust accuracy in \cref{tab:robust_overfit} is measured with AutoAttack robust accuracy and the clean accuracy is measured with clean data.

\subsection{The Effect of Temperature for Adversarial Distillation}

\begin{table}
    \centering
    {
    \begin{tabular}{lcccc}
    \toprule
     Temperature & 1 & 2  & 5 & 10  \\
    \toprule
     AKD$^2$ &26.11 &26.37 &26.44&26.22 \\
     RSLAD & 26.32& 24.70& 24.69 & 24.17\\
     IAD &25.60 & 25.94 &25.26&24.80 \\
     AdaAD & 24.89 & 24.37& 24.06 &24.22 \\
     \textbf{\thiswork} & \textbf{27.06} & \textbf{27.41}& \textbf{27.33} &\textbf{27.41} \\
    \bottomrule
    \end{tabular}
    } 
 \caption{AutoAttack robust accuracy of ResNet-18 trained by \thiswork and the baselines on CIFAR-100 under various temperature settings.}\label{tab:temp_sensi}
\end{table}
In \cref{tab:temp_sensi}, we tested the sensitivity of \thiswork and the baselines with respect to the temperature parameter of the loss in the student model.
Both \thiswork and the baselines have a distillation term, and we varied the temperature in the distillation term for the sensitivity study.
The adversarial distillation is conducted with ResNet-18 on CIFAR-100.
All other experimental settings are the same as in \cref{exp_setting}.
We tested four temperature values $\{1, 2, 5, 10\}$ and checked that \thiswork maintains higher AutoAttack robust accuracy in all tested temperatures.
This result shows that the higher robustness of \thiswork is insensitive to the the temperature parameter of the distillation term in the loss of the student model.

\subsection{Analysis of Training Time}

\begin{table}
    \centering
    \resizebox{1.0\columnwidth}{!}{
  \setlength{\tabcolsep}{.3\tabcolsep}
    \begin{tabular}{l|cc|c}
    \toprule
    Method & Pretraining Time & Distillation Time & Total Training Time\\
    \toprule
      Natural & - & -  & 2.01 hours (A) \\
      TRADES  & - & -  & 11.64 hours (B) \\
      \cmidrule(lr){1-4}
      AKD$^{2}$ & 13.65 hours (A+B) & 15.87 hours  & 29.52 hours  \\
      RSLAD  & 11.64 hours (B) & 26.21 hours & 37.85 hours \\
      IAD  & 11.64 hours (B) & 20.54 hours  & 32.18 hours  \\
      CAT  & - & 59.44 hours  & 59.44 hours  \\
      AdaAD  & 11.64 hours (B) & 28.89 hours  & 40.53 hours  \\
      \bf{\thiswork} & - & 30.50 hours & 30.50 hours  \\
    
    \bottomrule
    \end{tabular}
    } 
    \caption{Time cost of adversarial distillation methods with WideResNet34-10 on CIFAR-100.}\label{tab:training_time}
\end{table}

We checked the training time of \thiswork and the baselines to compare the computational cost among various methods.
As illustrated in \cref{tab:training_time}, \thiswork shows comparable total training time compared to other adversarial distillation methods.
Most of the baselines require a pretrained robust model except for CAT \cite{cat}.
We included the pretraining time in the total training time of the baselines, which required pretraining because the pretraining should be conducted beforehand to run them.
CAT requires nearly 2$\times$ time to perform adversarial distillation compared to \thiswork, though CAT is also an online adversarial distillation method.
The substantial computational cost of CAT arises from the necessity to attack two student models using different attack methods, whereas \thiswork only needs to attack one student model and \thiswork does not attack the peer model.
The time cost is measured with a single A100 GPU.

\clearpage

{
    \small
}
\end{document}